%% file: main.tex
\documentclass[11pt]{article}

\usepackage{microtype} 
\usepackage{booktabs}  
\usepackage{url}  
\usepackage{csquotes}

\usepackage{amsmath}
\usepackage{amsthm}

\usepackage{graphicx}
\usepackage{wrapfig}
\usepackage{caption}
\usepackage{natbib}

\input{commands.tex}

\usepackage[preprint]{automl}

\usepackage{natbib}

\newcommand{\submissiontitle}{%
  Searching on a Budget: HW-NAS with 10 Latency Probes
}

\title{\submissiontitle}
\author[1]{\nameemail{Francesco Capuano}{capuano@robots.ox.ac.uk}}
\author[3,4]{\nameemail{Gabriele Tiboni}{gabriele.tiboni@uni-wuerzburg.de}}
\author[2]{\nameemail{Niccolò Cavagnero}{niccolo.cavagnero@polito.it
}}
\author[2]{\nameemail{Giuseppe Averta}{giuseppe.averta@polito.it}}

\affil[1]{University of Oxford, Oxford, United Kingdom}
\affil[2]{Politecnico di Torino, Turin, Italy}
\affil[3]{Julius-Maximilians-Universität Würzburg, Würzburg, Germany}
\affil[4]{Technische Universität Darmstadt, Darmstadt, Germany}
\hypersetup{%
  pdfauthor={Francesco Capuano}, 
  pdftitle={\submissiontitle},
  pdfsubject={RL for NAS},
  pdfkeywords={AutoML, Reinforcement Learning, NAS}
}

\begin{document}

\maketitle

\begin{abstract}
    \input{sections/00_abstract.tex}
\end{abstract}

\section{Introduction}
\input{sections/01_introduction}

\section{Background}
\label{sec:background}
\input{sections/02_background}

\section{Method}
\input{sections/03_method}

\section{Experiments}
\input{sections/04_experiments}

\section{Conclusions}
\input{sections/05_conclusions}

\bibliographystyle{plainnat}
\bibliography{references}

\end{document}

%% file: commands.tex
\newcommand{\netpool}{\vert \mathcal N \vert}
\newcommand{\pfreerea}{p_\text{FreeREA}}
\newcommand{\phardware}{\bar{\ell}}
\newcommand{\dtarget}{d_\text{target}}

%% file: sections/00_abstract.tex
Hardware-aware NAS (HW-NAS) typically relies on analytical latency approximations or learned latency predictors, whose estimation errors can be detrimental in risk-sensitive applications. 
We propose a two-stage HW-NAS framework that first meta-trains an architecture controller across a distribution of synthetic, fictional devices, then deploys it on a target device using only direct latency readings. 
To scale meta-training, the controller is supervised with training-free accuracy proxies, avoiding the overhead of full network training. 
We benchmark on HW-NATS-Bench using data from 18 different devices, and validate the full deployment loop on a real-world, held-out commercial hardware platform.
Our pre-trained controller adapts to unseen hardware with as few as 10 latency measurements at test time. Under the same measurement budget, our method outperforms noisy latency predictors by designing networks with 8.7\% lower measured median latency.

%% file: sections/01_introduction.tex
Designing Deep Neural Networks (DNNs) for deployment on resource-constrained devices involves the two---oftetimes conflicting---objectives of \emph{(i)} selecting architectures that perform well, while \emph{(ii)} respecting hardware constraints.
\emph{Hardware-aware} Neural Architecture Search (HW-NAS) addresses such challenges automating architecture design under hardware supervision, offering a programmatic alternative to the otherwise-manual design of architectures that are performant \emph{and} efficient when deployed~\citep{li2021hw}.
However, most HW-NAS methods are typically tailored to a single target deployment, which in turn results in needing to re-search an architecture satisfying \emph{(i-ii)} whenever the target deployment---e.g., the \emph{device} \( d \)---changes.
We propose addressing these limitations by focusing on \emph{multi-device} HW-NAS.
In this, our goal is to discover architectures performing well on \emph{(i)} task-specific metrics---e.g., \emph{validation accuracy}---\emph{(ii)} as well as hardware-related metrics---e.g., \emph{post-compilation latency}---\emph{across} multiple, diverse devices.

Current HW-NAS approaches jointly tackling \emph{(i-ii)} typically rely on real-world performance measurements collected for a fixed target device~\citep{cai2019once,lee2021hardware,benmeziane2023efficient}, which can prove rather expensive to collect, as they may require compiling and benchmarking on-device a possibly large number of candidate networks.
Evaluating candidate designs can be more efficient by relying on \emph{(i)} analytical approximations of downstream hardware efficiency or \emph{(ii)} learning predictors approximating real-world measurements~\citep{laube2022expect}.
However, both of these approaches limit the applicability of HW-NAS in risk-averse scenarios---e.g., on-edge deployment for robotics---as \emph{(i)} relies on assumptions that often prove overly simplistic to hold in practice, while \emph{(ii)} is hindered by limited predictive accuracy and the difficulty of bounding the uncertainty of its estimates.

In this work, we propose a two-stage approach to HW-NAS mitigating the need to access information regarding the target device pre-deployment. 
Instead, we focus on developing a search-strategy aiming at performing HW-design for \emph{simulated, fictional} versions of the target device, to then transfer to real-world devices, optimizing the design of the network on edge by collecting few real-world probes, for test-time adaptation only.
Crucially, we directly and only use real-world latency measurements, thus completely avoiding inaccurate latency estimations.
In this, we find that pre-training on synthetic devices drastically limits the number of networks probed at test-time, making direct, real-world measurements a viable alternative, when learning to only probe \emph{the right networks} at test-time.
Indeed, by training on simulated devices \(d_i \sim \{d_1, \dots, d_n \}  \), the controller acquires a model of how latency/accuracy patterns vary across devices.
When deployed on a target device \(d \), our controller is capable of transfering its knowledge in order to \emph{more effectively explore} the tradeoffs between task performance and the (unknown) target latency distribution tradeoff.
This strategy allows to completely abandon \emph{amortized} latency estimations via approximations, or predictors, and instead allow to directly compile and measure real-world latency values on a handful of candidate networks.

Our contributions are summarized as follows:
\begin{itemize}
  \item \textbf{We develop a randomized RL training procedure for HW-NAS} that produces a controller able to interact directly with diverse devices at test time.
  \item \textbf{We entirely disregard analytical approximations and learned predictors of post-compilation latency}, in favor of direct on-device measurements on as few as 10 candidate network. Such result is made possible by the prior knowledge transferred from training on simulated devices.
  \item \textbf{We adopt training-free metrics for RL-based NAS}, leveraging cheap proxies to score candidate architectures, enabling rapid candidate evaluation without the cost of full training.
\end{itemize}

%% file: sections/02_background.tex
\begin{figure}
    \centering
    \includegraphics[width=0.99\linewidth]{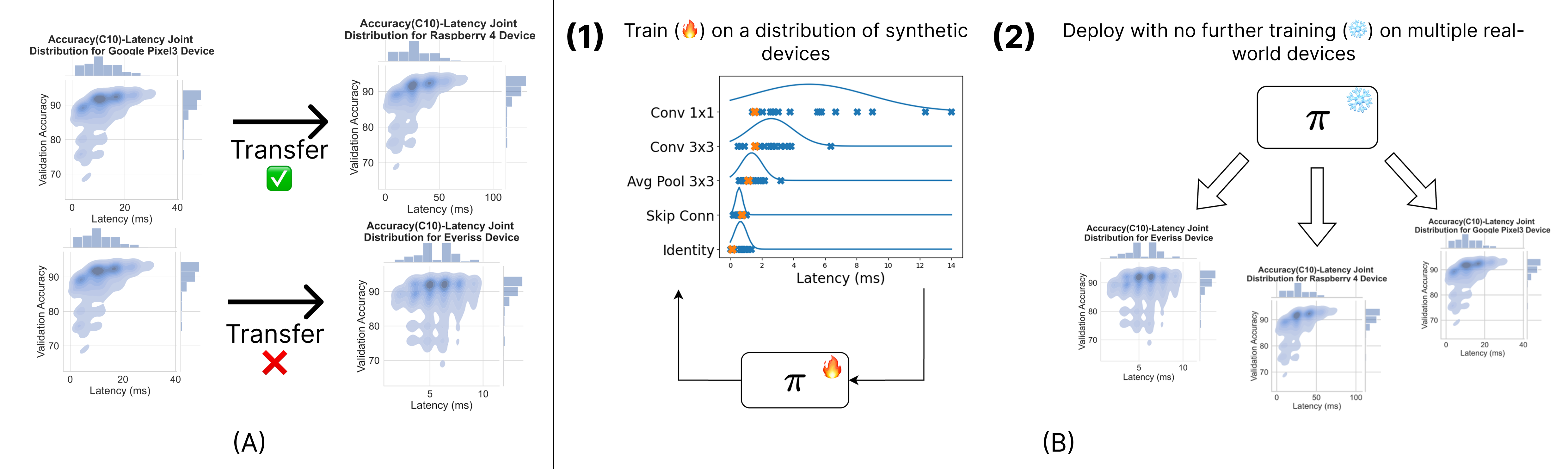}
    \caption{Overview of our method. (A) HW-NAS across different hardware platforms is hindered by fundamental differences across devices, which in turn results in different performance/efficiency tradeoffs across different devices. (B) Our method consists of a two-stage process where we first learn on a distribution of synthetic devices (B, 1), and then zero-shot transfer our learned policy for deployment on mulitple devices (B, 2).}
    \label{fig:fig1}
\end{figure}

\label{background}
\subsection{Hardware-Aware Neural Architecture Search (HW-NAS)}

Neural Architecture Search (NAS) automates DNN design by optimizing over a search space \( \mathcal H \), but early RL-based NAS required training many candidates and was therefore computationally prohibitive~\citep{zoph2016neural,zoph2018learning}. Later work reduced this burden through weight sharing~\citep{brock2017smash}, gradient-based search~\citep{liu2018darts}, and \emph{training-free} (TF) metrics that score architectures at initialization~\citep{mellor2021neural,jacot2018neural,chen2021neural,chen2023understanding}. Following~\citet{cavagnero2023freerea}, we use the efficient TF score \(\pfreerea(h)=\text{NASWOT}(h)+\text{LogSynflow}(h)+\text{SkipScore}(h)\), which combines expressivity, gradient-flow, and skip-connection signals while avoiding expensive training. Hardware-aware NAS (HW-NAS) extends NAS by optimizing architectures for deployment metrics such as latency, memory, or energy~\citep{benmeziane2023efficient,wu2019fbnet,king2025micronas,speckhard2022neural}; here we focus on post-compilation latency, which is critical for resource-constrained settings such as robotics. Measuring latency directly is expensive because each candidate may require device-specific compilation and timing, and measurements vary across hardware pipelines, compilers, and operator libraries~\citep{li2021hw,benmeziane2023efficient,laube2022expect}. Benchmarks such as HW-NAS-Bench~\citep{li2021hw} provide measured latency and energy for NATS-Bench architectures~\citep{dong2021nats}, but many HW-NAS methods still rely on approximations: lookup tables aggregate per-operation runtimes, \(\ell_d(a) \approx \hat{\ell}_d(a)=\sum_{o\in a}t_d(o)\)~\citep{wu2019fbnet,cai2018proxylessnas}, while learned predictors regress latency from architecture encodings using measured samples~\citep{laube2022expect}. LUTs are cheap but miss cross-operation effects, and predictors can outperform LUTs leveraging \emph{a few hundred real-world samples}, but introduce approximation risk and require upfront measurements~\citep{laube2022expect}. 

\paragraph{Low-cost HW-NAS}
If we model the total NAS cost as \( C=\sum_i^{\mathcal N} c_i \approx \netpool c \), where \(\netpool\) is the number of probed architectures and \(c\) is the cost per probe, low-cost HW-NAS methods limit \( C \) by bounding the total number of candidates evaluated \( \netpool \), the \emph{marginal} cost to score one candidate design \( c\), or both.
TF metrics and a pre-trained controller can be used to restrict \(\netpool\) to a small set of candidates that can be compiled and measured directly, while supernets, LUTs, or latency predictors allow to test a large number of networks by limiting \(c\).

Once-for-All (OFA)~\citep{cai2019once} amortizes NAS cost by training a large super-network once, then selecting child subnetworks whose weights are inherited without retraining. 
For HW-NAS, OFA uses LUT-guided EAs~\citep{real2017large} to find subnetworks satisfying device-specific constraints, which makes test-time evaluation cheap. 
However, the super-network pre-training phase itself is expensive---\citet{cai2019once} report 1.2k GPU hours on NVIDIA V100s---and therefore assumes resources unavailable in many low-budget settings. 
Further, its hardware-aware selection step also inherits LUT approximation errors, which can produce sub-optimal latency-accuracy tradeoffs~\citep{laube2022expect}.

Hardware-adaptive Efficient Latency Predictor (HELP)~\citep{lee2021hardware}---the work that most closely aligns with ours---proposes limiting \( C \), by learning a \emph{meta-predictor} to estimate post-compilation latency for unseen devices, focusing on \( c \downarrow \implies C \downarrow \).
Rather than measuring the latency of every candidate architecture on new physical devices, HELP learns a device-conditioned meta-predictor \( f: \mathcal Z \times \mathcal D \mapsto \mathbb R^+ \), mapping architecture encodings \( z \in \mathcal Z \) to hardware costs for different devices \( d \in \mathcal D \). 
Then, by assessing as little as 10 architectures on \(\dtarget \), HELP only adapts the meta-learned predictor, and limits the \emph{per-probe cost} \(c\) of each architecture.
Constructing HELP's predictor entails collecting a set of real-world measured samples \(\tau_d=\{h(a_i), \ell(a_i)\}_{i=1}^K\) across many different devices, which can prove challenging for non-experts in hardware~\citep{li2021hw}. 
Lastly, in risk-sensitive deployments even modest misestimation of latency can invalidate guarantees on latency bounds, hindering applicability of predictor-based approaches such as HELP. 

\paragraph{Ours}
Both OFA and HELP aim to mitigate \( C \) by limiting the overhead associated with \(c \). 
OFA amortizes \( c \) by reusing super-network weights and LUT---at the expense of accuracy in latency estimation---while HELP uses a meta-learned latency regressor---at the expense of generalization risks in latency estimation.
In contrast, we propose a method that explicitly mitigates \( C \) by limiting the \emph{number} of ground-truth probes \(\netpool \) probed at test-time, sidestepping super-network pre-training or predictive models altogether. 
This design choice inherently avoids approximation errors in hardware metrics, as each of the few final architectures is tested with real measurements. Hence, we shift complexity into a pre-trained RL controller (Section~\ref{sec:method}), trained to generalize over device variations in order to design well-performing DNNs across various target platforms.

\subsection{Reinforcement Learning (RL)}
Reinforcement learning (RL) is an area of machine learning concerned with training agents to act optimally in an environment that provides feedback via a reward signal~\citep{sutton1998reinforcement}. 
In general, RL agents interact with an environment based on its state \(s_t \in \mathcal S \) by taking an action \(a_t \in \mathcal A \), to then observe a next state \( s_{t+1} \) and reward \( r_t \).
Over the course of their interactions with the environment, RL agents attempt to learn a conditional probability distribution \(\pi: \mathcal S \times \mathcal A \mapsto [0, 1] \)---\emph{policy}, \(\pi(a \vert s)\)---following the objective of cumulative reward maximization.

Over discrete state or action spaces---typical in cell-based NAS where any \( s \in \mathcal S \) may represent a given cell/architecture \( h \in \mathcal H \), while actions \( a \in \mathcal A \) are synonomical with discrete modifications of \( h \)---RL has known a series of successes, starting with the seminal work of~\citet{mnih2013playing}.

While RL-based NAS set an important precedent, it was quickly recognized that repeatedly training every candidate network rendered the search protocol too costly, particularly considering RL's sample inefficiency~\citep{zoph2016neural,real2017large}.
Thus, researchers moved towards alternative solutions attempting to reduce computational demand by limiting the number of networks probed \( \netpool \), such as by using evolutionary algorithms (EAs)~\citep{real2017large,cavagnero2023freerea}.
While EAs have shown promise for single-device NAS~\citep{real2017large,cavagnero2023freerea}, we argue they remain ill-suited for multi-device NAS, or when training and target device differ, due to the challenges of transfering EAs' results across different HW-NAS instances.
In contrast, RL proved promising in fields where cross domain adaptability is key, such as robotics~\citep{kober2013reinforcement,akkaya2019solving}, and we thus argue for RL's return to HW-NAS, for its \emph{transfer learning} capabilities.
Crucially, cross-domain adaptability via RL could result in single controllers, usable across diverse devices.

%% file: sections/03_method.tex
\label{sec:method}

\begin{figure}
    \centering
    \includegraphics[width=0.8\linewidth]{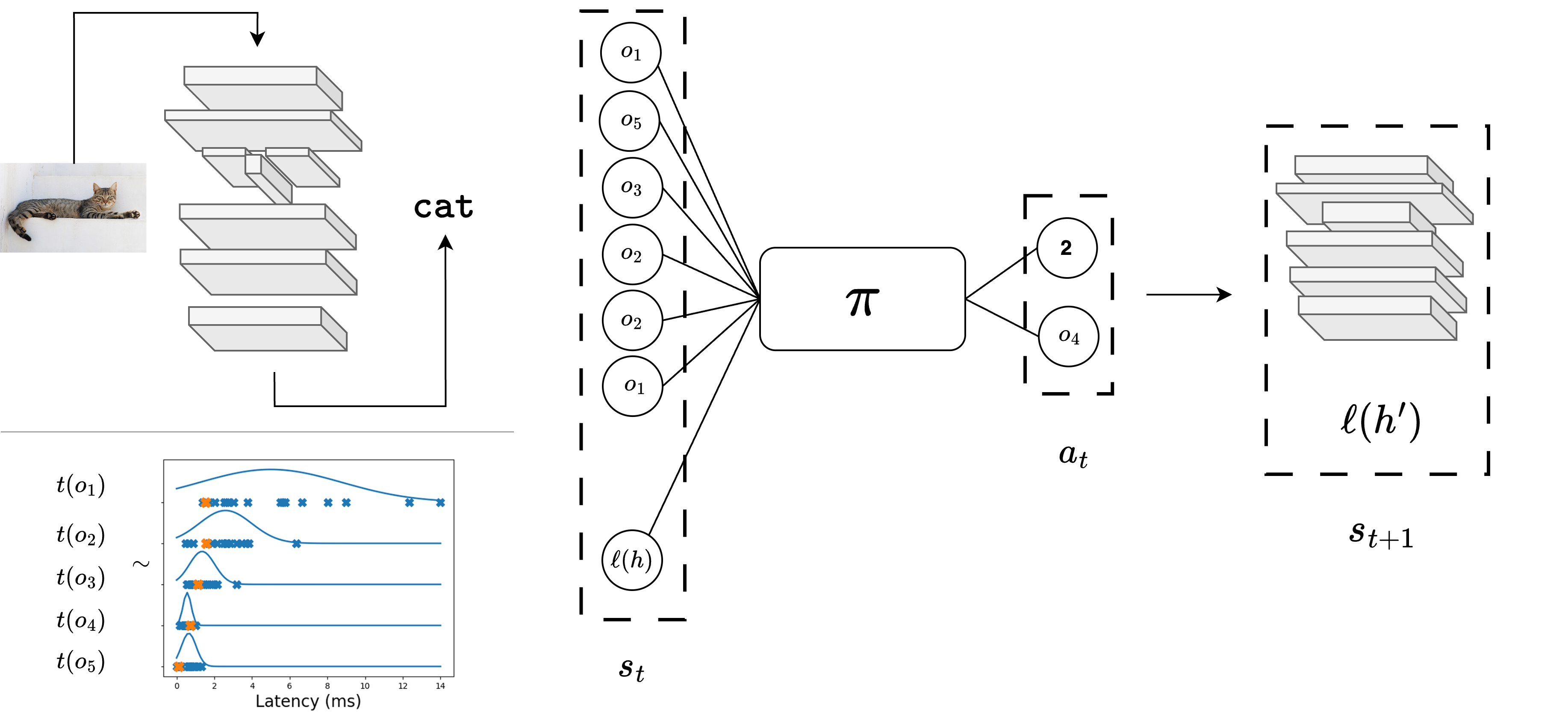}
    \caption{Overview of the policy network for our method. At training time, the policy \( \pi \) accesses (1) a candidate network, \( h \) and its associated latency \(\text{Latency}(h) \). The policy is then trained to propose a modification to \( h \) through modifying one of the operations in one of the positions. Over the course of training, latencies for the different networks are sampled from a distribution, resulting in synthetic devices.}
    \label{fig:policynet}
\end{figure}

We formulate low-budget HW-NAS as selecting an architecture \(h \in \mathcal H\) that preserves task performance while satisfying the latency profile of a target device \(d\):
\[
    \underset{h \in \mathcal H}{\max}\ \mathrm{Acc}(h)
    \quad \text{s.t.} \quad
    \mathrm{Latency}_{d}(h) \leq t_{\max}.
\]
We instantiate \(\mathcal H\) with the NATS-Bench~\citep{dong2021nats} topological search space and replace validation accuracy during search with the training-free proxy \(\pfreerea\). A candidate cell is represented as \(h=(o_1,\dots,o_6)\), with \(o_i \in \mathcal O_\text{NATS}\), and an action \(a=(i,o_\text{new})\) replaces operation \(o_i\). The controller receives the current architecture and its normalized latency on the active device, \(s_t=[h_t \Vert \ell(h_t)]\), and outputs the next edit. Rewards combine proxy quality and hardware efficiency,
\[
    r_\alpha(h)=\alpha \pfreerea(h)+(1-\alpha)\phardware(h),
    \qquad
    \phardware(h)=1-\ell(h),\quad \ell(h)\in[0,1],
\]
where \(\alpha\) sets the accuracy--latency operating point.

\paragraph{Training on randomized devices.}
To train a policy that transfers across hardware (Figure~\ref{fig:policynet}), we expose it to synthetic latency profiles instead of a single device. From the operation-wise measurements for 18 devices in HW-NAS-Bench~\citep{li2021hw,lee2021hardware}, we fit for each operation \(o \in \mathcal O_\text{NATS}\) a Gaussian latency model
\(
\mathcal N_o(\hat\mu_o^\mathcal D,(\hat\sigma_o^\mathcal D)^2).
\)
At the start of each episode, we sample a \emph{randomized} lookup table \(\xi=[t(o)]_{o\in\mathcal O_\text{NATS}}\), keep it fixed for that episode, and use it to compute \(\ell(h)\) and therefore \(r_\alpha(h)\). These sampled LUTs are not intended to predict any deployment device. Instead, they provide useful for \emph{domain randomization}~\citep{tobin2017domain,akkaya2019solving}, so that the policy learns generally plausible edits that remain useful under changing latency landscapes.

\paragraph{Deployment.}
At test time, the policy is frozen and deployed on a target device \(\dtarget\). 
It queries only a small number of architectures, measures their post-compilation latency directly on \(\dtarget\), appends the normalized measurement to the state history, and chooses the next edit conditioned on that feedback. 
The final architecture is selected from the measured candidates, avoiding target-device latency predictors and any architecture training during search. 
In our experiments, scoring the architectures within NATS-Bench with TF-scores such as \( \pfreerea \) runs in \( \sim 1 \) GPU hour, on a single NVIDIA RTX 4080.

%% file: sections/04_experiments.tex
\label{sec:experiments}

Our experiments evaluate whether the controller described in Section~\ref{sec:method} can adapt to unseen target devices using only a small number of true target-device latency measurements.
Indeed, our method explicitly targets a deployment where latency measurements on \( \dtarget \) are accurate but scarce, while downstream accuracy labels and target-device latency predictors are unavailable during search.

Using the cost decomposition \(C \approx \netpool c\) from Section~\ref{sec:background}, our method effectively reduces deployment cost by reducing \(\netpool\), i.e. the number of architectures probed on the target device at runtime.
This contrasts with predictor-based hardware-aware NAS methods such as HELP~\citep{lee2021hardware}, which reduce the effective \emph{per-candidate cost} \(c\) by replacing target-device measurements with latency predictions.
Thus, our experiments compare two deployment strategies: direct low-budget adaptation through measured latency feedback, and predictor-based search through calibrated target-device latency models, and aim at addressing three key questions:
\begin{enumerate}
    \item[Q1] Can our pre-trained controller transfer to held-out devices using only a small target-device measurement budget?
    \item[Q2] How does this direct-measurement strategy compare with HELP-style latency prediction under the same scarce-measurement setting?
    \item[Q3] Does the latency history collected on the target device condition the search trajectory, yielding device-specific architectures?
\end{enumerate}

\paragraph{Experimental setup}
We evaluate on the NATS-Bench topological search space~\citep{dong2021nats} for CIFAR-10.
The search space contains 15,625 candidate cells constructed from the operation set \( \mathcal O_\text{NATS} \).
Following~\citet{cavagnero2023freerea}, each architecture is assigned the training-free performance proxy \( \pfreerea \).
This TF proxy is the only task-performance signal used during search, as validation and test accuracies from NATS-Bench are used only \emph{after} search has completed, as reporting metrics.
Target-device latency is either obtained from the measured latency tables provided by HW-NAS-Bench~\citep{li2021hw}, or measured directly on a real-world device (iPhone 13).

As our method aims at reducing \( C \) limiting \( \netpool \), we mainly compare against HELP-style latency prediction~\citep{lee2021hardware}, which in turn reduce \( C \) by limiting \( c \) directly.
Our HELP-style baseline uses a small number (20) of target-device measurements to calibrate a latency predictor, which is then integrated with an Evolutionary Algorithm to guide hardware-aware search (HELP-EA).
In practice, the HELP predictor uses source-device latency profiles as meta-features, calibrates a ridge latency predictor from a small set of target-device probes, and ranks candidate architectures by predicted latency.
Crucially, OFA~\citep{cai2019once} is not included as a direct baseline because it assumes a separately trained super-network, whereas our setting assumes no target-task training during search, nor low-cost HW-NAS \emph{via amortization}.

\paragraph{Training and deployment protocol}
Our controller is trained using the randomized-LUT procedure described in Section~\ref{sec:method}.
For each held-out target device \(d\), we use a leave-one-device-out protocol: all latency measurements from \(d\) are removed from the device pool used to construct the training-time latency distributions, and \(d\) is used only for deployment-time evaluation.
This ensures that transfer is evaluated on devices that were not observed during meta-training.
We train PPO~\citep{schulman2017proximal} for 500k environment steps, using a discount factor of \(\gamma=0.6\), a learning rate \(3\times 10^{-4}\), and the default PPO clipping setup.
Episodes start from a randomly sampled NATS-Bench architecture and last \(T=50\) steps during training, resulting in 50 architectures probed for episode (at most).
Unless otherwise stated, the reward uses equal weights (\( \alpha = 0.5 \)) for the training-free performance proxy and hardware-efficiency terms.

At test time, the trained policy is frozen and deployed on a held-out target device \(d\).
No gradient update, latency-predictor fitting, or architecture training is performed during deployment.
The controller is allowed to query at most 10 candidate architectures on \(d\), thereby limiting test episodes to a duration of 10 timesteps in total.
After each query, the measured latency is appended to the policy history and can influence the next action. 
We repeat evaluation over 10 random initial architectures and seeds, and report both the terminal architecture and the best measured architecture found within the query budget.
The latter is the practically relevant deployment choice: since every queried architecture has a measured target-device latency and a known training-free proxy value, selecting the best candidate among the measured set requires no additional information.

\paragraph{Real-world validation}

\begin{wrapfigure}{r}{0.28\linewidth}
    \centering
    \includegraphics[width=\linewidth]{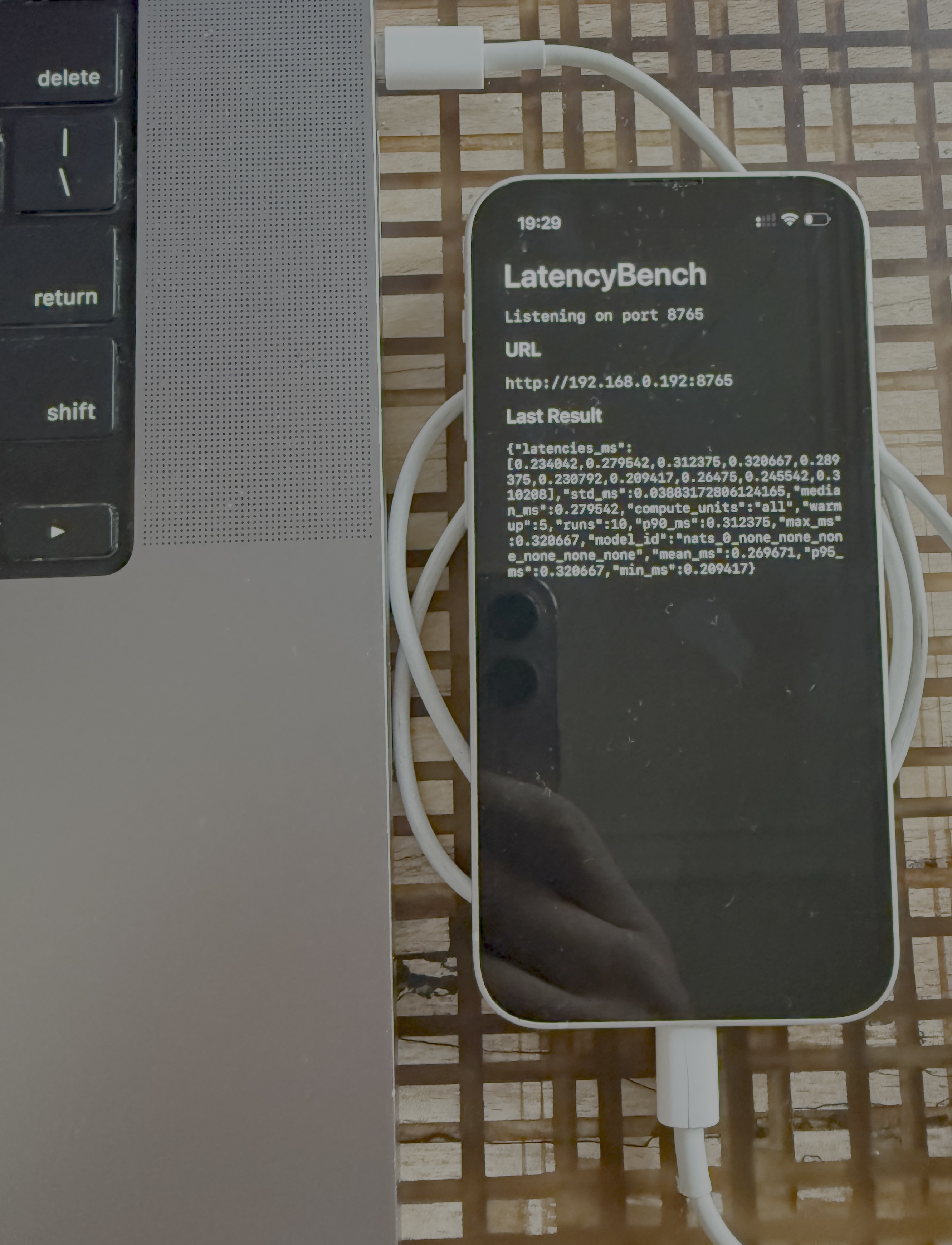}
    \caption{Real-world deployment on an iPhone 13-mini.}
    \label{fig:iphone_deployment}
\end{wrapfigure}

We additionally validate the deployment loop on a real iPhone 13-mini (Figure~\ref{fig:iphone_deployment}).
In practice, each candidate architecture is exported from PyTorch to Core ML, and benchmarked on hardware using a thin iOS benchmarking app, further undescoring the development challenges and costs arising from doing research considering hardware as well.
The app compiles the uploaded \texttt{.mlmodel} with Core ML, constructs a fixed zero-valued CIFAR-shaped input tensor, performs warmup predictions, and then times repeated calls to \texttt{MLModel.prediction}, a hardware utility exposed by Core ML~\citep{apple_coreml}.
Upload, compilation, and model-loading time are negligible, and thus excluded from the reported latency.

Before deployment, we measure 10 calibration architectures on the iPhone and compute robust bounds \(q_{0.05}^\text{cal}\) and \(q_{0.95}^\text{cal}\) to obtain \( \tilde{\ell} \), using the 5th and 95th percentiles of their measured \(\log L_{\dtarget}(h)\).
Each subsequent physical latency measurement is converted to the controller's latency input by
\(
    \tilde{\ell}_{\dtarget}(h_t)
    =
    \operatorname{clip}\left(
        \frac{\log L_{\dtarget}(h_t) - q_{0.05}^\text{cal}}
        {q_{0.95}^\text{cal} - q_{0.05}^\text{cal}},
        0, 1
    \right).
\)
This transformation maps physical latency units into the scale seen during meta-training.
For a fair comparison, these 10 calibration probes are included in the physical-device measurement budget, while raw physical latency remains the reported deployment metric and is used to audit the selected architecture.

In our real-world experiment, methods receive the same 20 pre-decision iPhone measurements, but allocate them differently.
Our controller uses 10 measurements to fit the normalizer and 10 additional measurements for adaptive direct search.
On the other hand, HELP-EA uses the same 10 normalizer probes plus 10 additional calibration probes to fit its latency predictor, then searches by predicted tradeoff only.
The HELP-EA selected architecture is measured on the iPhone after selection solely to audit the predictor-based decision.
This audit measurement is not used by the HELP-EA search procedure.

\begin{table}
    \centering
    \caption{Deployment-time comparison under a 10-query target-device measurement budget. Values are means \(\pm\) standard deviations over test episodes. Lower latency percentile is better. HELP-style methods reduce search cost by lowering per-candidate evaluation cost \(c\) through prediction; our controller reduces search cost by bounding the number of true target-device probes \(\netpool\).}
    \label{tab:help_comparison}
    \small
    \setlength{\tabcolsep}{3.2pt}
    \begin{tabular}{llcccc}
        \toprule
        Method & Devices & Lever & Lat. pct. \(\downarrow\) & Test acc. \(\uparrow\) & Tradeoff \(\uparrow\) \\
        \midrule
        Random search & 4 & \(\netpool\downarrow\) & \(30.16 \pm 21.59\) & \(87.63 \pm 11.48\) & \(0.665 \pm 0.049\) \\
        \(\pfreerea\)-top-\(K\) & 4 & \(\netpool\downarrow\) & \(84.82 \pm 20.20\) & \(94.05 \pm 0.18\) & \(0.653 \pm 0.045\) \\
        HELP-style & 4 & \(c\downarrow\) & \(4.51 \pm 6.09\) & \(87.21 \pm 1.83\) & \(\mathbf{0.788 \pm 0.027}\) \\
        Ours & 4 & \(\netpool\downarrow\) & \(\mathbf{1.65 \pm 4.68}\) & \(85.50 \pm 4.14\) & \(0.734 \pm 0.024\) \\
        \midrule
        Official HELP & 3 & \(c\downarrow\) & \(11.00 \pm 4.01\) & \(\mathbf{90.59 \pm 0.00}\) & \(\mathbf{0.775 \pm 0.011}\) \\
        Ours & 3 & \(\netpool\downarrow\) & \(\mathbf{4.40 \pm 11.68}\) & \(86.00 \pm 3.07\) & \(0.738 \pm 0.032\) \\
        \bottomrule
    \end{tabular}
\end{table}

\begin{table}
    \centering
    \caption{Physical iPhone 13 mini validation over 10 seeds. Both methods receive 20 pre-decision iPhone measurements. Ours allocates them to 10 normalization probes and 10 adaptive direct-search probes; HELP-EA allocates them to predictor calibration and then selects by predicted tradeoff only. The HELP-EA audit is measured only after selection and is not used by the predictor-based search.}
    \label{tab:iphone_validation}
    \begin{tabular}{lp{0.24\textwidth}ccc}
        \toprule
        Method & Measurements & Med. lat. (ms) \(\downarrow\) & Tradeoff \(\uparrow\) \\
        \midrule
        HELP-EA & 20 & \(0.2477 \pm 0.0134\) & \(0.5614 \pm 0.1359\) \\
        \textbf{Ours} & 20 & \(\mathbf{0.2261 \pm 0.0058}\) & \(\mathbf{0.7010 \pm 0.0800}\) \\
        \bottomrule
    \end{tabular}
\end{table}

\begin{table}
    \centering
    \caption{Latency diagnostics for HELP-EA on the same iPhone 13 mini runs. Predicted latency is the value used by HELP-EA to select the final architecture; measured latency is obtained only with post-selection audit.}
    \label{tab:iphone_latency_diagnostics}
    \begin{tabular}{p{0.42\textwidth}ccp{0.24\textwidth}}
        \toprule
        Quantity & Mean \(\pm\) std. & Seeds & Interpretation \\
        \midrule
        Predicted latency (ms) & \(0.2263 \pm 0.0085\) & 10 & used for selection \\
        Measured latency (ms) & \(0.2477 \pm 0.0134\) & 10 & post-selection audit \\
        Error (ms), measured-predicted & \(0.0214 \pm 0.0140\) & 10 & \emph{slower} than predicted \\
        \bottomrule
    \end{tabular}
\end{table}

\begin{figure}
    \begin{minipage}{0.60\textwidth}
        \centering
        \captionsetup{width=0.95\textwidth}
        \includegraphics[width=0.8\linewidth]{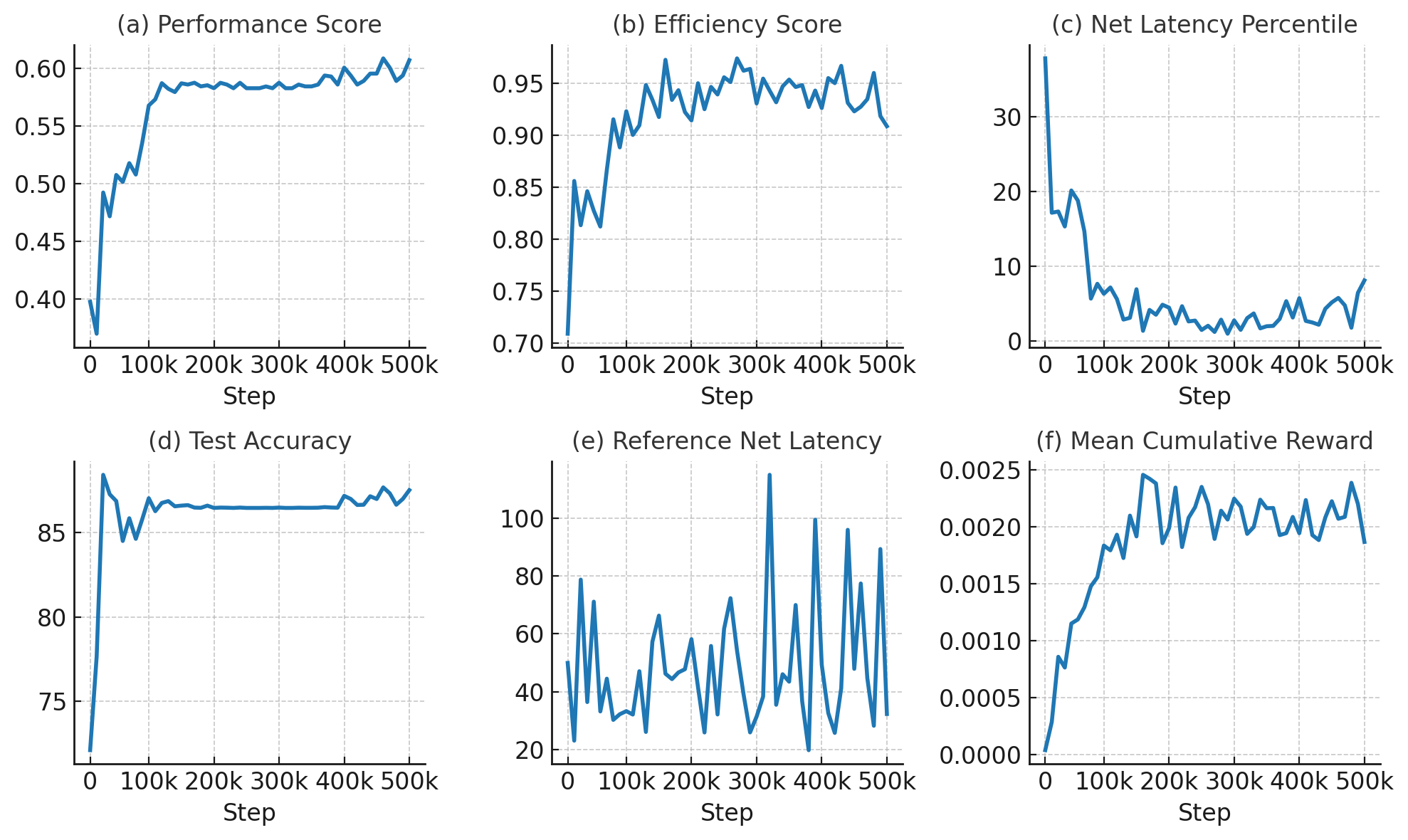}
        \caption{Average results collected over 20 test episodes during training. (a) Average normalized \( \pfreerea(h_T)\). (b) Average normalized \(\phardware(h_T)\). (c) Average latency percentile of \( h_T \). (d) Average final validation accuracy of \( h_T \), never accessed during training. (e) Average latency of a fixed reference network \( h_{\text{ref.}} \). (f) Average cumulative reward over training.}
        \label{fig:trainingruns}
    \end{minipage}
    \hfill
    \begin{minipage}{0.38\textwidth}
        \centering
        \captionsetup{width=0.95\textwidth}
        \includegraphics[width=0.8\textwidth]{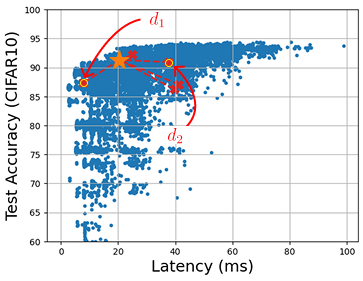}
        \caption{Starting from the same network (orange star), the policy reaches different terminal configurations for different target devices (red crosses).}
        \label{fig:deviceroutes}
    \end{minipage}
\end{figure}

\paragraph{Q1: Transfer under a small target-device measurement budget}
Table~\ref{tab:help_comparison} evaluates whether the pre-trained controller can adapt to unseen devices using only a small number of true target-device latency measurements.
The 4-device block holds out \texttt{edgegpu}, \texttt{fpga}, \texttt{pixel3}, and \texttt{raspi4} one at a time; for each held-out target, we run 3 random seeds. The Official HELP block uses the three targets covered by the released HELP assets~\citep{lee2021hardware}: \texttt{fpga}, \texttt{pixel3}, and \texttt{raspi4}.
Across the four-device leave-one-out evaluation, our controller reaches an average best latency percentile of \(1.65\) after 10 target-device measurements.
This improves over random search, which reaches \(30.16\), and over the HELP-style predictor baseline, which reaches \(4.51\) under the same target-device measurement budget.
On the subset of devices supported by the official HELP NAS-Bench-201 assets, the same pattern holds: our controller reaches a lower average latency percentile than the official HELP checkpoint used as a target-adapted predictor, finding faster networks, though at the (marginal) expense of accuracy.

\paragraph{Q2: Direct low-budget adaptation versus latency prediction}
The comparison with HELP-style methods exposes the intended cost-quality tradeoff.
HELP reduces deployment cost by learning a latency predictor from a small calibration set and then ranking many unmeasured candidate architectures cheaply.
Our controller instead keeps latency evaluation high-fidelity and measured, but reduces the number of architectures that must be evaluated on the target device.
Accordingly, the two approaches optimize different deployment regimes.

Our method is designed for a regime where deployment decisions should rely on a very small number of true target-device measurements and no target-latency surrogate.
In that regime, our controller reaches lower measured latency percentiles, at the cost of a less accuracy-oriented operating point.
Different performance-latency tradeoffs can in principle be obtained by varying the reward weights.

The physical iPhone 13 mini experiment in Table~\ref{tab:iphone_validation} further validates this distinction on a real target device.
Under the same 20 pre-decision iPhone measurements, our controller finds architectures with \(0.2261 \pm 0.0058\)ms median latency, compared with \(0.2477 \pm 0.0134\)ms for HELP-EA.
This corresponds to an \(8.7\%\) lower measured median latency.
The measured scalar tradeoff is also higher for our controller, \(0.7010 \pm 0.0800\) versus \(0.5614 \pm 0.1359\).

Table~\ref{tab:iphone_latency_diagnostics} shows why this physical-device comparison is informative.
Across 10 iPhone 13 mini runs, HELP-EA's latency predictor failed to match deployment measurements: predicted and measured latency differed in all seeds, and measured latency was slower in 9/10 cases. As a result, HELP-EA was worse than our method in measured latency for 10/10 seeds and in accuracy-latency tradeoff for 8/10 seeds.
Thus, under the same physical measurement budget, the predictor-based search often selected an architecture whose realized latency was worse than expected.
By contrast, our controller selects among architectures whose target-device latency has already been measured.

\paragraph{Training dynamics and proxy behavior}
Figure~\ref{fig:trainingruns} shows that the controller improves both components of the training objective over the course of synthetic-device training.
The increase in normalized \( \pfreerea \) indicates that the policy learns to move toward architectures preferred by the training-free proxy.
At the same time, the improvement in \(\phardware\) and latency percentile shows that this behavior is not achieved by ignoring hardware efficiency.
The validation-accuracy, reference network's latency and reward panels are exclusively used as an external check. Indeed, test-accuracy stability shows that proxy-driven optimization does not simply select low-latency architectures with poor downstream task quality, while the reference network latency variability provides information on the level of diversity induced by randomized LUTs. The reward evolution is used to verify the agent succesfull learns to solve it task.
All values are averaged over 20 rollout episodes periodically run during training.

\paragraph{Q3: Device-conditioned search trajectories}
Figure~\ref{fig:deviceroutes} evaluates whether the target latency history changes the policy trajectory.
Architectures are plotted with respect to the same latency profile for the sake of visualization.
Starting from the same architecture, the controller follows different routes under different target-device latency measurements and terminates in different regions of the search space.
This behavior is the intended consequence of the history-based observation: the same trained policy is reused across devices, but the measured latency sequence changes the state history and therefore the selected edits.

%% file: sections/05_conclusions.tex
\label{sec:conclusion}

In this work, we presented a low-budget formulation of multi-device HW-NAS in which a controller is trained across randomized synthetic device profiles and then deployed on unseen hardware using only direct latency measurements. By combining training-free performance proxies with a device-conditioned RL policy, our method avoids both super-network pretraining and target-device latency prediction, and instead reduces search cost by bounding the number of target-device probes.

Empirically, this strategy transfers across held-out HW-NAS-Bench devices under a 10-query deployment budget, and it also remains effective on a real iPhone 13 mini. In that physical deployment, our controller achieves an \(8.7\%\) lower median measured latency than the predictor-based baseline under the same pre-decision measurement budget. These results support the main claim of the paper: when access to the target hardware is scarce but each measurement must be trustworthy, it can be preferable to measure a few carefully chosen architectures rather than predict many unmeasured ones.

\paragraph{Limitations}
Due to limited research in HW-NAS, this study is limited to the NATS-Bench topology search space, so it remains to be shown how well the approach scales to larger search spaces and more realistic training pipelines. 
Extending the method to settings such as FBNet~\citep{wu2019fbnet}, or to broader hardware-aware benchmarks, is therefore a natural next step.
Nonetheless, we believe the experiments and real-world validation of our method to still prove effective tests.

A second limitation is that the training-time synthetic device family is deliberately simple
While randomized LUTs are sufficient to induce useful transfer in our experiments, the effect of the design of \( \Xi : \xi \sim \Xi \) on downstream adaptation remains underexplored. 
In particular, too little diversity may hinder transfer, while overly broad randomization may over-regularize the controller and weaken device-specific adaptation. Better procedures for constructing or scheduling synthetic device distributions, potentially inspired by curriculum-based domain randomization~\citep{akkaya2019solving,tiboni2023domain}, are an important direction for future work.

Finally, our experiments focus primarily on latency and on a low-measurement deployment regime. Practical deployment often requires balancing multiple hardware objectives, including accuracy, latency, memory, and energy. Extending the controller to richer multi-objective operating regimes, while preserving the core principle of direct low-budget measurement on the target device, is in our view the most important next step toward broadly usable HW-NAS. Still, we believe bounding latency with a small number of trustworthy on-device measurements to the most impactful first step toward real-world deployment of networks discovered with HW-NAS.